\documentclass[letterpaper]{article} 
\usepackage[preprint]{utils/aaai2027}  
\usepackage[hyphens]{url}  
\usepackage{graphicx} 
\usepackage{amsmath}
\usepackage{natbib}  
\AtBeginDocument{\setcitestyle{numbers,square,sort&compress}}
\usepackage{caption} 
\usepackage{algorithm}
\usepackage{algorithmic}

\usepackage{newfloat}
\usepackage{listings}
\DeclareCaptionStyle{ruled}{labelfont=normalfont,labelsep=colon,strut=off} 
\floatstyle{ruled}
\newfloat{listing}{tb}{lst}{}
\floatname{listing}{Listing}

\usepackage{booktabs}
\usepackage{array}
\usepackage{pifont}

\definecolor{supportedpink}{HTML}{E91E63}
\definecolor{partialgold}{HTML}{FFE600}
\definecolor{absentcyan}{HTML}{00A6B4}
\newcommand{\capyes}{\textcolor{supportedpink}{\ding{51}}}
\newcommand{\cappart}{\textcolor{partialgold}{\ding{108}}}
\newcommand{\capno}{\textcolor{absentcyan}{\ding{55}}}

\title{GenPuzzle: Benchmarking Visual Reasoning in Image Generation Models}
\author{
    Changpeng Zhao\textsuperscript{\rm 1},
    Yiren Song\textsuperscript{\rm 2},
    Jinpeng Wang\textsuperscript{\rm 1}\corresponding
}
\affiliations{
    \textsuperscript{\rm 1}Central South University\\
    \textsuperscript{\rm 2}National University of Singapore
}

\makeatletter
\let\genpuzzle@original@maketitle\@maketitle
\renewcommand{\@maketitle}{%
  \genpuzzle@original@maketitle
  \centering
  \includegraphics[width=0.98\textwidth]{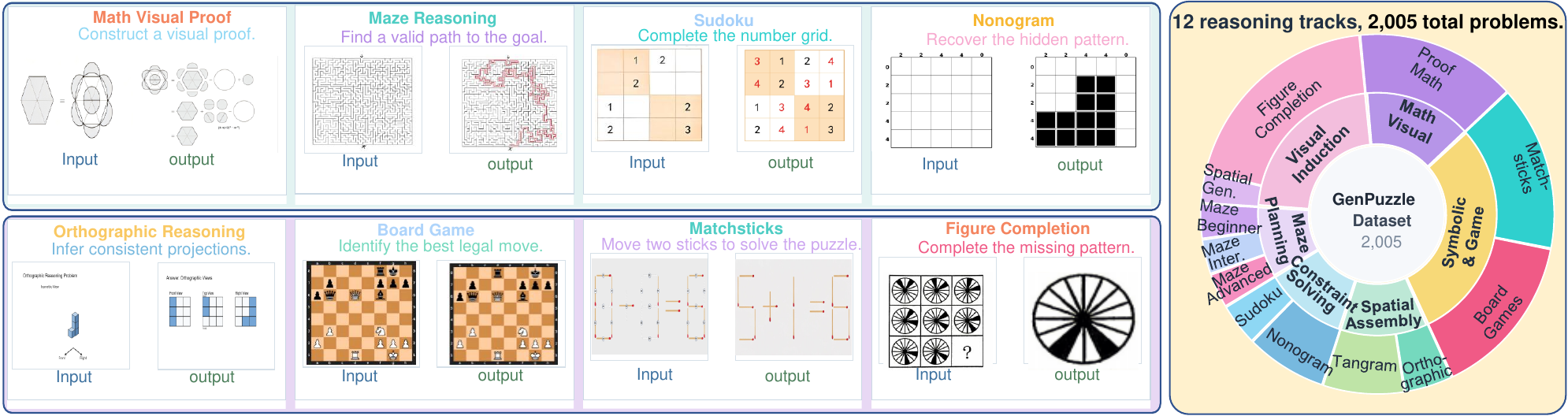}\par
  \captionof{figure}{Overview of GenPuzzle. Given a puzzle image and instruction, an image generation model produces a visual solution that is assessed using a task-specific evaluation protocol. The benchmark contains 2,005 problems across 12 tracks; representative input--output pairs and the track distribution are shown.}
  \label{fig:teaser}
}
\makeatother

\begin{document}

\maketitle

\begin{abstract}
Recent image generation systems increasingly combine multimodal understanding, reasoning, and synthesis, suggesting that they may do more than render plausible scenes. 
Yet existing evaluations emphasize aesthetics, prompt alignment, compositionality, or text-based answers, leaving unclear whether these systems can solve visual problems and faithfully express solutions in pixels. 
We introduce GenPuzzle, a benchmark for reasoning-centric image generation. 
GenPuzzle contains 2,005 problems across 12 tracks, spanning pattern completion, spatial construction, mazes, Sudoku, nonograms, tangrams, board games, matchstick puzzles, orthographic projection, and mathematical visual proof. 
Each task provides a visual puzzle and requires an image output that preserves the input state while executing a logically valid solution. 
GenPuzzle uses task-specific evaluation protocols: discrete grid outputs are transcribed and verified programmatically, while visually complex outputs are assessed with tiered, multidimensional, or binary multimodal large language model (MLLM) rubrics. 
We further select the automatic judge by measuring agreement with human reference scores. 
Across three frontier generators, the strongest model reaches only 40.57 Macro Overall, revealing frequent failures in logic, geometry, state preservation, and instruction execution. 
GenPuzzle provides a testbed for measuring progress from image rendering toward visual problem solving.
\end{abstract}

\begin{links}
    \link{Project}{https://github.com/zhaochangpeng-csu/GenPuzzle}
\end{links}

\section{Introduction}

Modern visual generation models have evolved from text-conditioned renderers into multimodal systems that can interpret reference images, follow complex instructions, and produce structured visual outputs~\cite{wiedemer2025video,li2025girbench,jiang2025t2ir1}.
This progress raises a question beyond image quality and prompt adherence: \emph{Can image generation models reason through images?} In this paper, reasoning means that a model understands a visual problem state, infers a valid solution, and realizes that solution in pixels without corrupting the unchanged parts of the input.

Table~\ref{tab:benchmark_comparison} compares representative
benchmarks along the capabilities required for executable
visual problem solving, including direct generation evaluation,
image-based inputs and outputs, rule-governed correctness,
state preservation, alternative valid solutions, and
constructive visual answers.

\begin{table}[!t]
\centering
{\footnotesize
\setlength{\tabcolsep}{1.7pt}
\renewcommand{\arraystretch}{1.05}
\begin{tabular}{@{}>{\raggedright\arraybackslash}p{0.31\columnwidth}ccccccc@{}}
\toprule
\textbf{Benchmark} & \textbf{DIG} & \textbf{In} & \textbf{Out} & \textbf{Rule} & \textbf{State} & \textbf{Open} & \textbf{Constr.} \\
\midrule
DrawBench~\cite{saharia2022imagen} & \capyes & \capno & \capyes & \cappart & \capno & \capno & \capno \\
T2I-CompBench~\cite{huang2023t2icompbench} & \capyes & \capno & \capyes & \cappart & \capno & \capno & \capno \\
GenAI-Bench~\cite{li2024genaibench} & \capyes & \capno & \capyes & \cappart & \capno & \capno & \capno \\
MMMU~\cite{yue2024mmmu} & \capno & \capyes & \capno & \cappart & \capno & \capno & \capno \\
BLINK~\cite{fu2024blink} & \capno & \capyes & \capno & \cappart & \capno & \capno & \capno \\
MathVerse~\cite{zhang2024mathverse} & \capno & \capyes & \capno & \capyes & \capno & \capno & \capno \\
RBench-V~\cite{guo2025rbenchv} & \cappart & \capyes & \cappart & \capyes & \cappart & \capno & \capyes \\
RISEBench~\cite{zhao2025risebench} & \cappart & \capyes & \capyes & \capyes & \capyes & \capno & \cappart \\
GIR-Bench~\cite{li2025girbench} & \capyes & \cappart & \capyes & \capyes & \cappart & \capno & \cappart \\
BabyVision-Gen~\cite{chen2026babyvision} & \capyes & \capyes & \capyes & \cappart & \capyes & \capno & \cappart \\
\textbf{GenPuzzle} & \capyes & \capyes & \capyes & \capyes & \capyes & \capyes & \capyes \\
\bottomrule
\end{tabular}
}
\caption{Capability comparison. DIG: direct generation eval.; In/Out: image input/output answer; Rule: explicit rule- or constraint-based correctness; State: input preservation; Open: alternative valid solutions; Constr.: constructive visual answers. \capyes: supported; \cappart: partial; \capno: absent.}
\label{tab:benchmark_comparison}
\end{table}

Existing image generation benchmarks mainly measure perceptual quality, text--image alignment, object counting, attribute binding, and compositional consistency~\cite{saharia2022imagen,lee2023heim,ghosh2023geneval,huang2023t2icompbench,li2024genaibench}. These tasks reveal important rendering failures, but they usually describe the desired image directly. A model can therefore succeed by drawing what is stated, rather than deriving an unstated solution. Conversely, multimodal reasoning benchmarks such as MMMU, BLINK, MathVerse, and VisuLogic evaluate visual reasoning but usually ask for a textual answer or a selected option~\cite{yue2024mmmu,fu2024blink,zhang2024mathverse,xu2025visulogic}. Recent BabyVision-Gen~\cite{chen2026babyvision} extends visual reasoning evaluation
to generative outputs by adapting foundational perceptual
tasks into minimally annotated image responses
for 280 generation-based questions. However, it primarily focuses on early
visual competencies and annotation-style solutions. Broader
rule-governed visual problem solving---including discrete
constraint satisfaction, strategic actions, precise state
transformations, and tasks admitting alternative valid
solutions---remains comparatively underexplored. GenPuzzle
complements this line of work by requiring models to preserve
a concrete input state, infer a valid solution under
task-specific rules, and execute that solution directly in
image space.

Recent work has started to connect reasoning with multimodal outputs, visual editing, and reasoning-sensitive synthesis~\cite{guo2025rbenchv,zhao2025risebench,li2025girbench,jin2024reasonpix2pix,he2025reason50k,yin2025reasonedit,jiang2025t2ir1,duan2025gotr1,zhang2025reasongenr1}. However, broad evaluation of frontier image generators on image-conditioned puzzle solving remains limited. This setting is demanding because correctness has two inseparable parts: the inferred solution must obey the task rules, and the generated image must execute it faithfully. A model may know the answer but draw it incorrectly, or produce a plausible image that silently changes the puzzle state.

We formulate this problem as \emph{visual reasoning through generation}. The model receives a puzzle image and an instruction, infers a valid solution under the task rules, and returns an image that executes the solution. Unlike visual question answering, the answer is not a word, coordinate, or option. It must be carried out in the visual domain, such as drawing a legal maze path, filling a Sudoku grid, applying a legal board-game move, or moving exactly the permitted matchsticks. The output must also preserve all input elements unrelated to the required operation, a consistency requirement that remains central to controllable visual generation~\cite{song2025omniconsistency}. This makes the task a joint test of reasoning, visual editing, and state preservation.

We introduce \textbf{GenPuzzle}, a benchmark for this capability. It contains 2,005 problems across 12 equally weighted tracks: visual pattern completion, spatial construction, three maze difficulties, Sudoku, nonograms, tangrams, board-game reasoning, matchstick manipulation, orthographic projection, and mathematical visual proof. These tracks cover discrete and continuous spaces, local and global transformations, deterministic and open-ended solutions, and two- and three-dimensional reasoning. Figure~\ref{fig:teaser} summarizes the workflow and task distribution.

Evaluating such outputs also requires task-aware protocols. Pixel similarity penalizes alternative valid solutions, while a generic multimodal judge may miss violations such as an illegal move, a wall-crossing path, or a deformed tangram piece. GenPuzzle therefore combines structured transcription and programmatic verification for grid-based tasks with tiered, multidimensional, and binary MLLM rubrics for visually complex tasks. Candidate judges are selected by agreement with human reference scores.

Our experiments show that strong visual synthesis does not imply reliable visual reasoning. Under the selected Gemini 3.1 Pro judge, Seedream 5.0 Pro, Nano Banana 2.0, and GPT Image 2 obtain Macro Overall scores of 40.57, 32.37, and 29.41, consistent with human ranking. Failures are especially severe on matchstick puzzles and mazes, where models must preserve a precise state while performing constrained local or geometric operations.

Our contributions are summarized as follows:
\begin{itemize}
\item We introduce \textbf{GenPuzzle}, a benchmark that evaluates visual reasoning through image generation across 2,005 problems and 12 track-balanced evaluation categories.
\item We construct a structured benchmark with task-specific annotations, reference solutions, open-solution policies, and evaluation metadata for diverse puzzle families.
\item We design validated task-specific evaluation protocols and benchmark frontier generators, revealing a large gap between visually polished outputs and logically correct visual solutions.

\end{itemize}

\FloatBarrier

\section{Related Work}
\subsection{Image Generation and Evaluation}

Recent image generation models have achieved remarkable progress in visual quality, prompt following, and image-conditioned synthesis. Accordingly, existing benchmarks primarily evaluate whether generated images faithfully reflect explicitly specified textual concepts. DrawBench~\cite{saharia2022imagen} broadens human evaluation of text-to-image systems, TIFA~\cite{hu2023tifa} decomposes prompts into visual question-answer pairs to measure fine-grained faithfulness, and GenEval~\cite{ghosh2023geneval} evaluates object presence, counting, colors, positions, and attribute binding using object-centric detectors. T2I-CompBench~\cite{huang2023t2icompbench} and GenAI-Bench~\cite{li2024genaibench} further examine compositional generation involving attributes, spatial relations, relationships, and logic-like prompt constraints. Although these benchmarks effectively reveal failures in semantic alignment and compositionality, their prompts generally describe the expected visual outcome directly. Consequently, models can succeed through faithful rendering without deriving a latent solution. In contrast, our benchmark requires models to first solve an underspecified visual problem and subsequently execute the inferred solution in image space.

\subsection{Visual Reasoning Benchmarks}

Visual reasoning benchmarks evaluate whether multimodal models can integrate perception, knowledge, and logical inference. MMMU~\cite{yue2024mmmu} collects college-level multimodal questions across diverse disciplines, while MMMU-Pro~\cite{yue2024mmmupro} reduces language shortcuts by filtering text-solvable questions and introducing vision-only settings. BLINK~\cite{fu2024blink}, VisuLogic~\cite{xu2025visulogic}, and ZeroBench~\cite{roberts2025zerobench} emphasize core perception, abstract patterns, spatial relations, counting, and other tasks in which visual information is indispensable. Mathematical benchmarks such as MathVista~\cite{lu2024mathvista} and MathVerse~\cite{zhang2024mathverse} further probe diagram-based mathematical reasoning. RBench-V~\cite{guo2025rbenchv} additionally studies reasoning supported by multimodal intermediate outputs. Nevertheless, most existing benchmarks ultimately assess a textual response, such as a multiple-choice option, number, or explanation. They therefore primarily test whether models can extract an answer from an image, rather than whether models can preserve a visual state and accurately instantiate the solution through image generation.

\subsection{Reasoning-Centric Visual Generation}

Recent studies have begun to connect explicit reasoning with visual generation and editing. ReasonPix2Pix~\cite{jin2024reasonpix2pix} studies instruction reasoning for advanced image editing, while RISEBench~\cite{zhao2025risebench} evaluates image editing instructions involving temporal, causal, spatial, and logical reasoning. GIR-Bench~\cite{li2025girbench} studies understanding-generation consistency, reasoning-sensitive text-to-image synthesis, and multi-step visual editing. Broader controllable editing methods address local handle transformations, efficient image editing, concept swapping, and instruction-driven editing~\cite{ma2023magicstick,yan2025eedit,zhu2024instantswap,feng2025dit4edit,huang2025arteditor,zhang2025easycontrol}. Complementary approaches investigate controllable layouts, visual relation transfer, and structured vector synthesis~\cite{wan2024grid,gong2025relationadapter,song2025layertracer,song2023clipvg}. Meanwhile, methods such as ReasonEdit~\cite{yin2025reasonedit}, T2I-R1~\cite{jiang2025t2ir1}, GoT-R1~\cite{duan2025gotr1}, and ReasonGen-R1~\cite{zhang2025reasongenr1} introduce thinking, reflection, chain-of-thought supervision, or reinforcement learning to improve compositional and spatial generation. These efforts demonstrate growing interest in reasoning-aware synthesis, but largely focus on semantic constraint satisfaction, implicit-knowledge prompting, or general-purpose editing. GenPuzzle instead treats image generation as the answer interface for deterministic visual problems, jointly evaluating solution correctness, geometric execution, and preservation of the original puzzle state across diverse rule-governed tasks.

\section{Benchmark Construction and Evaluation}
\label{sec:method}

\subsection{Task Formulation}

We define visual reasoning in image generation as the ability to infer a solution from an input image and express that solution correctly in the generated output. The $i$-th benchmark instance is
\begin{equation}
q_i = \left(I_i, x_i, t_i, \mathcal{R}_i, M_i\right),
\label{eq:instance}
\end{equation}
where $I_i$ is the input puzzle image, $x_i$ is the natural-language instruction, $t_i$ is the task category, $\mathcal{R}_i$ is the set of task rules, and $M_i$ is the metadata used only for evaluation. A model produces $\hat{I}_i=G(I_i,x_i)$ with access only to the puzzle image and instruction. Reference solutions, structured ground truth, rubrics, and hidden metadata are withheld during generation.

A correct output must satisfy four requirements: logical validity, visual execution correctness, input-state preservation, and instruction compliance. These criteria distinguish reasoning errors from generation errors. For example, a model may infer a legal board-game move but redraw unrelated pieces, or preserve the board while executing an illegal move.

\subsection{Dataset and Benchmark Construction}

GenPuzzle contains 2,005 problems across 12 task tracks: visual pattern completion, spatial construction, beginner mazes, intermediate mazes, advanced mazes, Sudoku, nonograms, tangrams, board games, matchstick arithmetic puzzles, orthographic projection, and mathematical visual proof. The three maze difficulty levels are separate tracks because they differ in path length, branching structure, and planning complexity. Together, the tracks evaluate visual induction, path planning, discrete constraint satisfaction, strategic reasoning, precise manipulation, geometric composition, two- and three-dimensional spatial understanding, and visually grounded mathematical reasoning. The mathematical visual proof track complements vision-centric mathematical benchmarks such as MathVista~\cite{lu2024mathvista}.

GenPuzzle adopts a unified structured data format. Each instance contains a puzzle image, a task instruction, one or more reference solution images, and task-specific annotations used for evaluation. For paired-image tasks, including visual pattern completion, spatial construction, mazes, Sudoku, nonograms, and tangrams, the core annotations consist of the input image, reference solution image, and a unique task identifier. For more structured tasks, including board games, matchstick puzzles, orthographic projection, and mathematical visual proof, the annotations additionally include structured ground truth, such as board states, valid actions, target equations, voxel representations, projection views, proof objectives, and required visual constructions.

The benchmark combines publicly accessible puzzle sources with procedurally generated instances. It includes 394 visual pattern completion problems, 56 spatial construction problems, three maze tracks of 64 problems each, 78 $4\times4$ Sudoku puzzles, 150 nonograms, 150 tangrams, 300 board-game problems covering 20 games or rule-based board puzzles, 300 matchstick equation transformations, 90 orthographic-projection problems, and 295 mathematical visual proof problems. For open-solution tasks, the reference image is one known valid solution rather than the only acceptable answer.

We apply multi-stage quality control throughout data construction. Task-specific checks verify the uniqueness of Sudoku and nonogram solutions, the geometric validity of tangram configurations, the legality of board-game actions, the movement constraints of matchstick transformations, the projection consistency of orthographic-view problems, and the alignment between proof objectives and ground-truth diagrams in mathematical visual proof tasks. Finally, we conduct a manual audit to verify that each JSON record is correctly matched with its associated input and reference images.

\subsection{Task-Specific Evaluation}

The requirements for a correct output vary across tracks, so neither a single image-similarity metric nor a shared evaluation prompt is sufficient. This limitation is also observed in prior work on compositional, conditional, and diffusion-based visual similarity evaluation~\cite{ghosh2023geneval,ku2024viescore,song2024diffsim}, and motivates validating MLLM judges against human judgments and difficult error cases~\cite{lu2023llmscore,saxon2024t2iscorescore}. Each GenPuzzle protocol specifies evaluation inputs, scoring rubrics, fatal-error definitions, and policies for alternative valid solutions. Judges receive the original puzzle, candidate output, reference solution, task annotations, and rubric. All sample-level scores are mapped to $[0,100]$.

GenPuzzle uses four scoring mechanisms. First, visual pattern completion, spatial construction, mazes, and tangrams use task-specific four-tier MLLM rubrics with normalized score $s_i^{(t)}=r_i^{(t)}/3\times100$, where $r_i^{(t)}\in\{0,1,2,3\}$. These rubrics check rule following, path validity, geometric constraints, piece preservation, and alternative valid solutions; severe state corruption imposes task-specific score caps. Second, Sudoku and nonogram outputs are transcribed into discrete grids and verified programmatically. Sudoku verification checks given-digit preservation, blank-cell completion and accuracy, and row, column, and $2\times2$ subgrid constraints. Nonogram verification checks row and column clues, completion, and cell accuracy.

\begin{figure*}[t]
\centering
\includegraphics[width=\textwidth]{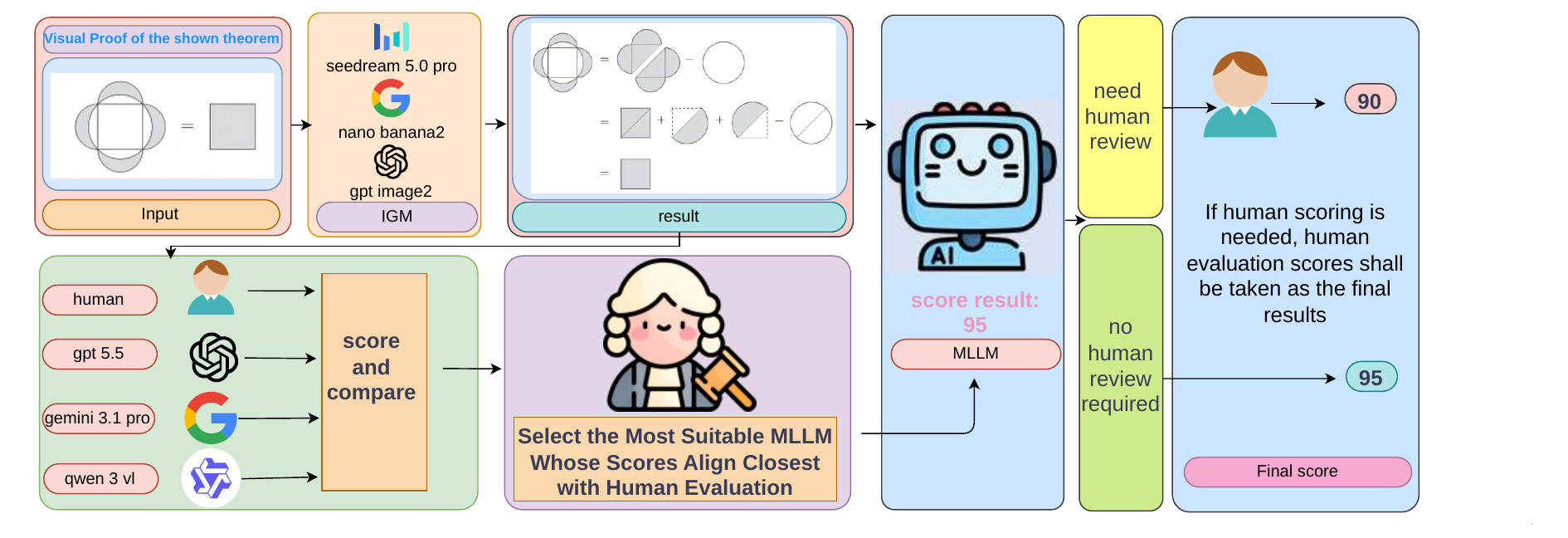}
\caption{Overview of the GenPuzzle generation and evaluation pipeline. The image generation model (IGM) receives only the puzzle image and instruction and produces a candidate visual solution. Candidate MLLM judges are compared against human reference scores; the selected judge then applies task-specific rubrics and flags uncertain outputs for human review.}
\label{fig:evaluation_pipeline}
\end{figure*}

Figure~\ref{fig:evaluation_pipeline} summarizes the end-to-end generation and evaluation workflow.

Third, board games, orthographic reasoning, and mathematical visual proof use task-specific five-dimensional MLLM rubrics with a total score of 100. For a task $t$ with $K_t$ scoring dimensions, the score of instance $i$ is
\begin{equation}
\begin{aligned}
s_i^{(t)}
&=
\sum_{k=1}^{K_t}s_{i,k}^{(t)}.
\end{aligned}
\label{eq:multidim_score}
\end{equation}
Board games are scored for layout, logical correctness, input preservation, text or label accuracy, and task completion, with weights of $15/45/20/10/10$. Orthographic reasoning uses weights of $20/35/20/15/10$ for instruction following, spatial and projection correctness, visual structure, label accuracy, and task completion. Mathematical visual proof uses weights of $20/25/30/15/10$ for instruction following, visual structure, reasoning correctness, text and symbol accuracy, and clarity. These text- and symbol-level criteria reflect the continuing challenge of controllable text rendering and typography in image generation~\cite{lu2025easytext,shi2024fonts,shi2025wordcon}. Illegal moves, inconsistent projections, incorrect conclusions, missing core constructions, or failure to use the input trigger fatal errors or score caps.

Fourth, matchstick puzzles use a strict binary rubric:
\begin{equation}
s_i^{\mathrm{matchstick}}
=
\begin{cases}
100, & \text{all constraints are satisfied},\\
0, & \text{otherwise}.
\end{cases}
\label{eq:matchstick_score}
\end{equation}

Full credit requires a valid final equation, legible digits and operators, conservation of the total number of matchsticks, and exactly the required number of moved sticks. Outputs that appear to be unannotated valid solutions, or whose critical visual elements are difficult to read, are flagged for human review.

\subsection{Score Aggregation}

Because GenPuzzle tracks contain different numbers of instances, we use the track-balanced Macro Overall score as the primary ranking metric~\cite{lee2023heim,yue2024mmmu}. Let track $t$ contain $N_t$ instances and let $s_{t,i}\in[0,100]$ be the normalized score of instance $i$. The score of track $t$ is
\begin{equation}
\begin{aligned}
S_t
&=
\frac{1}{N_t}
\sum_{i=1}^{N_t}s_{t,i}.
\end{aligned}
\label{eq:track_score}
\end{equation}

The Macro Overall score is defined as
\begin{equation}
\begin{aligned}
S_{\mathrm{macro}}
&=
\frac{1}{T}
\sum_{t=1}^{T}S_t,
\end{aligned}
\label{eq:macro_score}
\end{equation}
where $T=12$ for the complete benchmark. Each track contributes equally, with a weight of $1/12\approx8.33\%$, regardless of its number of instances.

\section{Experiments}
\label{sec:experiments}

\begin{figure*}[!t]
\centering
\includegraphics[width=\textwidth]{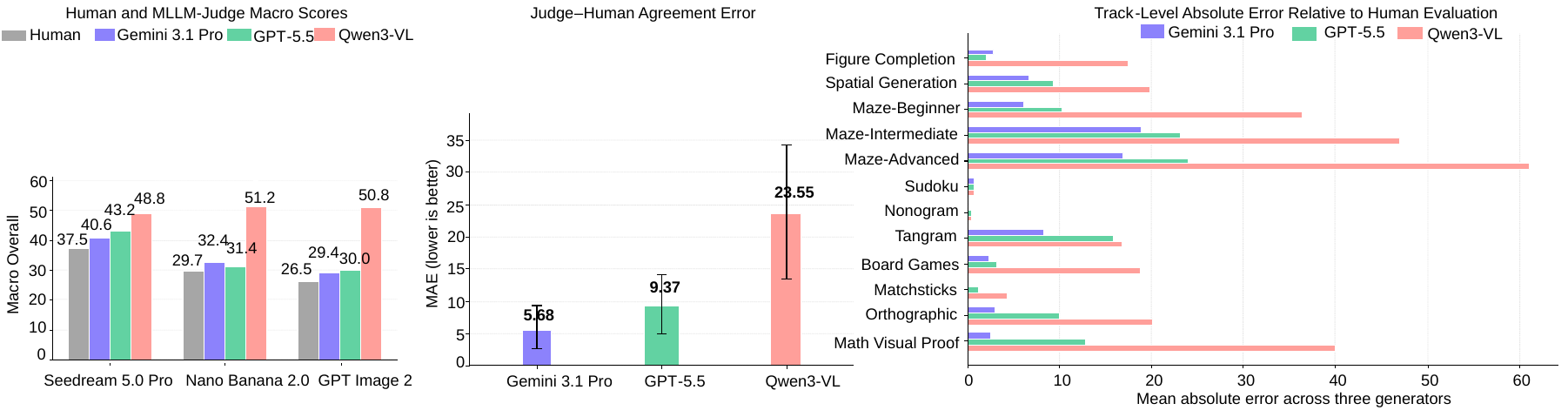}
\caption{Comparison of candidate judges against human reference evaluation. Left: Macro Overall scores assigned by humans and the three judges. Center: judge--human MAE with track-clustered bootstrap 95\% confidence intervals. Right: track-level mean absolute error relative to human evaluation.}
\label{fig:judge_selection}
\end{figure*}

\subsection{Experimental Setup}
\label{sec:experimental_setup}

Our experiments address two questions: which multimodal judge best agrees with human evaluation, and how frontier image generation models perform on GenPuzzle. We evaluate Seedream 5.0 Pro~\cite{bytedance2026seedream5pro}, Nano Banana 2.0, and GPT Image 2, generating one candidate image for each of the 2,005 instances. Generation models receive only the puzzle image and instruction. Candidate judges receive the puzzle image, generated candidate, reference solution, task annotations, and task-specific rubric, and then score the output using the benchmark-construction protocols above.

\subsection{Judge Selection}
\label{sec:judge_agreement}

Following prior work that validates automatic image evaluation against human judgments~\cite{ku2024viescore,saxon2024t2iscorescore}, we manually score the outputs of the three generators with the same task-specific criteria as the automatic judges. The human Macro Overall scores are 37.51, 29.69, and 26.47 for Seedream 5.0 Pro, Nano Banana 2.0, and GPT Image 2, respectively. To avoid letting larger tracks dominate judge selection, we compare judges on 36 generation-model--track aggregate points. Let $h_i$ denote the human score and $\hat{h}_{j,i}$ denote the score assigned by judge $j$. We use MAE as the primary agreement metric:
\begin{equation}
\operatorname{MAE}_j
=
\frac{1}{N}
\sum_{i=1}^{N}
\left|
\hat{h}_{j,i}-h_i
\right|.
\label{eq:judge_mae}
\end{equation}

We also report RMSE, bias, Pearson and Spearman correlations, and Macro-MAE, and estimate MAE uncertainty with 20,000 track-clustered bootstrap resamples. Absolute errors are compared with a Friedman test followed by paired Wilcoxon signed-rank tests with Holm correction.

\begin{center}
{\small
\setlength{\tabcolsep}{2.7pt}
\begin{tabular}{lccc}
\toprule
\textbf{Metric} & \textbf{Gemini 3.1 Pro} & \textbf{GPT-5.5} & \textbf{Qwen3-VL} \\
\midrule
MAE $\downarrow$ & \textbf{5.68} & 9.37 & 23.55 \\
95\% CI & \textbf{[2.61, 9.28]} & [4.97, 13.99] & [13.58, 34.19] \\
RMSE $\downarrow$ & \textbf{9.33} & 12.88 & 30.61 \\
Bias & +2.78 & +3.52 & +18.95 \\
Pearson $\uparrow$ & \textbf{0.912} & 0.826 & 0.435 \\
Spearman $\uparrow$ & \textbf{0.849} & 0.731 & 0.302 \\
Macro-MAE $\downarrow$ & \textbf{2.89} & 3.63 & 19.05 \\
\bottomrule
\end{tabular}
}
\captionof{table}{Agreement between candidate judges and human reference evaluation. Arrows indicate the preferred direction.}
\label{tab:judge_agreement}
\end{center}

Table~\ref{tab:judge_agreement} shows that Gemini 3.1 Pro has the lowest MAE (5.68) and RMSE (9.33), the highest Pearson (0.912) and Spearman (0.849) correlations, and the smallest Macro-MAE. GPT-5.5 preserves the broad ranking but has larger errors, while Qwen3-VL strongly overestimates performance. The Friedman test finds significant judge differences ($\chi^2=36.54$, $p=1.16\times10^{-8}$); after Holm correction, Gemini 3.1 Pro is significantly closer to humans than GPT-5.5 ($p=3.56\times10^{-4}$) and Qwen3-VL ($p=2.33\times10^{-5}$). These results suggest that an MLLM judge can support aggregate automatic evaluation, but it should not fully replace human evaluation because ambiguous visual cases still require manual review. We therefore select \textbf{Gemini 3.1 Pro}. Figure~\ref{fig:judge_selection} visualizes the same comparison.

\subsection{Image Generation Model Performance}
\label{sec:generation_performance}

After selecting Gemini 3.1 Pro, we freeze the judge and all evaluation settings. Table~\ref{tab:track_results} shows that Seedream 5.0 Pro obtains the highest Macro Overall score of 40.57, outperforming Nano Banana 2.0 and GPT Image 2 by 8.20 and 11.16 points. Nano Banana 2.0 ranks second with 32.37, while GPT Image 2 scores 29.41. This ranking is consistent with human evaluation.

\begin{figure*}[!t]
\centering
\includegraphics[width=\textwidth]{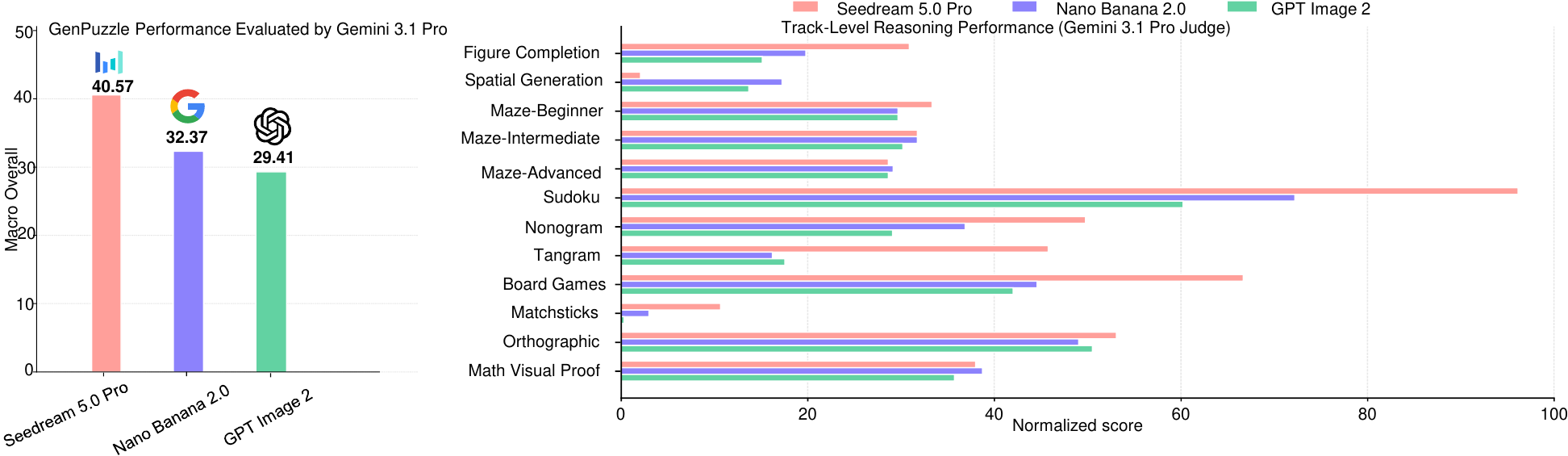}
\caption{Gemini 3.1 Pro evaluation of Seedream 5.0 Pro, Nano Banana 2.0, and GPT Image 2, with Macro Overall on the left and normalized GenPuzzle track scores on the right.}
\label{fig:generator_comparison}
\end{figure*}

Performance varies sharply by task. Seedream 5.0 Pro reaches 96.15 on Sudoku and leads on most tracks, but scores only 2.07 on spatial generation. Nano Banana 2.0 leads on spatial generation, advanced mazes, and mathematical visual proof, and ties Seedream 5.0 Pro on intermediate mazes. GPT Image 2 does not lead any track, although it is competitive on orthographic reasoning and mathematical visual proof.

The hardest cases expose the gap between rendering and reasoning. \textbf{The best matchstick score is only 10.67}, indicating failures in equation validity, matchstick-count conservation, and precise local editing. This low score is expected under the binary matchstick rubric: a solution must simultaneously read the input equation, move exactly the required sticks, conserve the total stick count, and render legible seven-segment symbols. Evaluation records frequently show arithmetically true equations that changed the stick count, alongside plausible local edits that moved the wrong number of sticks. Scores on the three maze tracks remain around 29--33, showing that models struggle to draw continuous wall-avoiding paths while preserving the original maze.

\begin{table}[!t]
\centering
{\small
\setlength{\tabcolsep}{2.6pt}
\begin{tabular}{lccc}
\toprule
\textbf{Track} & \textbf{Seedream} & \textbf{Nano Banana} & \textbf{GPT Image} \\
\midrule
Macro Overall & \textbf{40.57} & 32.37 & 29.41 \\
\midrule
Pattern Completion & \textbf{30.89} & 19.80 & 15.14 \\
Spatial Generation & 2.07 & \textbf{17.26} & 13.69 \\
Maze--Beginner & \textbf{33.33} & 29.68 & 29.68 \\
Maze--Intermediate & \textbf{31.77} & \textbf{31.77} & 30.21 \\
Maze--Advanced & 28.64 & \textbf{29.16} & 28.64 \\
Sudoku & \textbf{96.15} & 72.22 & 60.26 \\
Nonogram & \textbf{49.78} & 36.89 & 29.11 \\
Tangram & \textbf{45.78} & 16.22 & 17.55 \\
Board Games & \textbf{66.69} & 44.60 & 42.03 \\
Matchsticks & \textbf{10.67} & 3.00 & 0.33 \\
Orthographic & \textbf{53.09} & 49.04 & 50.51 \\
Math Visual Proof & 38.01 & \textbf{38.74} & 35.74 \\
\bottomrule
\end{tabular}
}
\caption{Overall and track-level scores from the selected Gemini 3.1 Pro judge. Best results are bolded.}
\label{tab:track_results}
\end{table}

Figure~\ref{fig:generator_comparison} combines the overall and track-level comparisons. It shows Seedream 5.0 Pro as the strongest aggregate model, while the track-level view exposes uneven profiles across all three generators, especially on tasks requiring exact local edits, state preservation, or constrained geometric execution.

\begin{figure*}[!t]
\centering
\includegraphics[width=\textwidth]{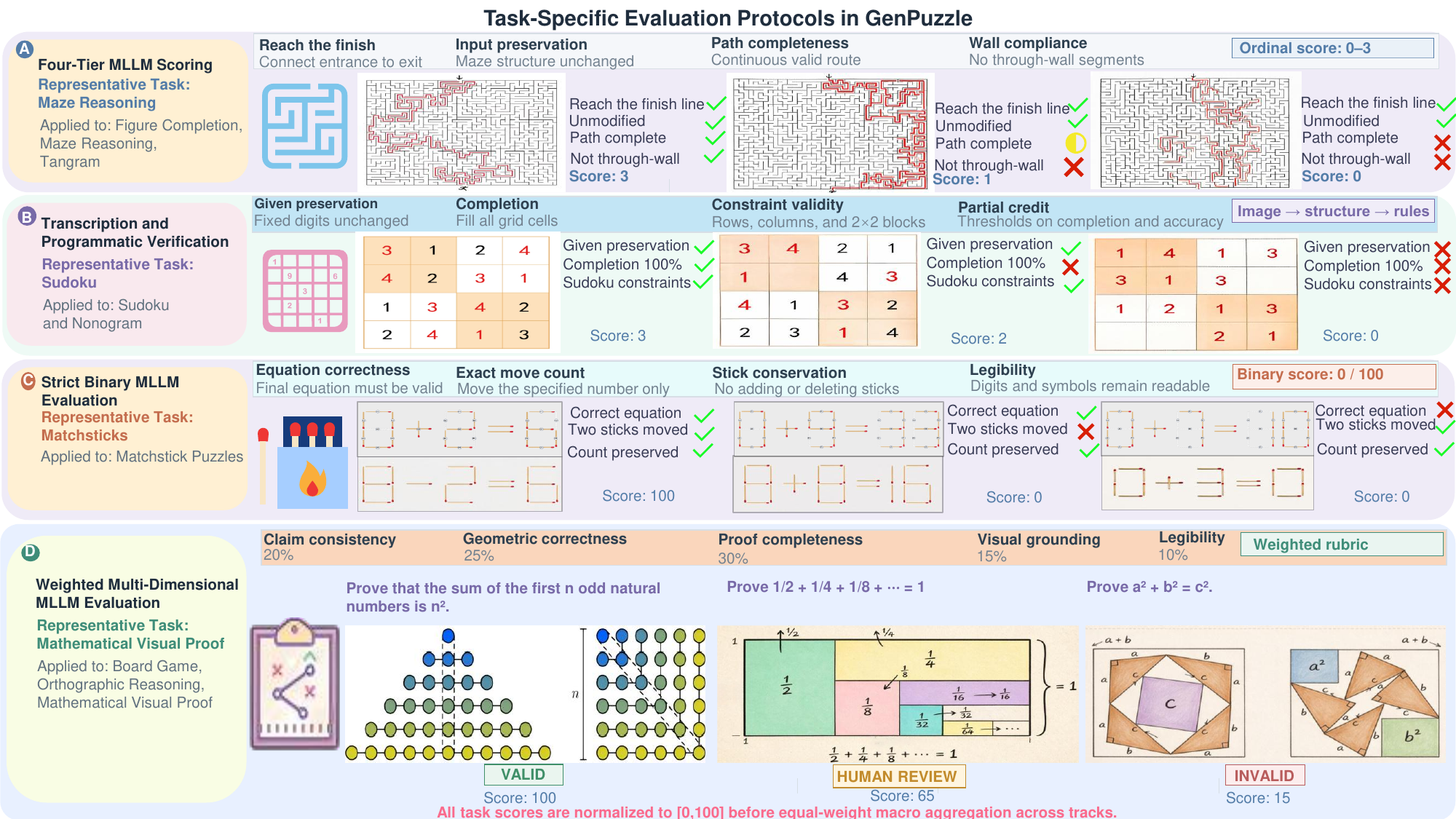}
\caption{Task-specific evaluation protocols in GenPuzzle: tiered MLLM scoring, transcription with programmatic checks, binary matchstick scoring, and weighted multidimensional rubrics with human review. Scores are normalized to $[0,100]$ before aggregation.}
\label{fig:evaluation_criteria}
\end{figure*}

\subsection{Qualitative Analysis}
\label{sec:qualitative_analysis}

Figure~\ref{fig:evaluation_criteria} illustrates how the evaluator distinguishes visual plausibility from task completion and logical correctness, complementing prior findings that difficult generation errors can expose weaknesses hidden by aggregate alignment metrics~\cite{saxon2024t2iscorescore}. Successful outputs must satisfy task-specific constraints, preserve relevant input content, and provide complete visual solutions. By contrast, polished images can fail because of incomplete proofs, discontinuous or wall-crossing paths, invalid Sudoku grids, or incorrect matchstick moves. Ambiguous cases are flagged for human review when key visual elements are difficult to read or insufficient to support a definitive judgment.

This design is especially important for generative answers, where a sample can be visually fluent yet logically unusable. Human review is therefore reserved for boundary cases rather than used as the default scoring mechanism, allowing GenPuzzle to scale while preserving a check on failure modes that are difficult to formalize.

The qualitative cases also show why the rubrics separate execution, preservation, and reasoning. In mazes, a visually salient red path can still cross a wall or stop before the goal. In Sudoku and nonograms, locally plausible digits or filled cells may violate global row and column constraints. In matchstick and proof tasks, the output must preserve small symbols while applying a precise transformation or construction. These failures are easy to miss with holistic image similarity, but they are central to visual problem solving. The evaluation protocol therefore treats a generated image as an executable answer, not only as a plausible picture. This framing makes error cases more diagnostic: a low score indicates which part of the pipeline failed, such as inference, state preservation, geometric execution, or legibility.

Together, these examples motivate reporting task-specific failure modes alongside aggregate scores.

\FloatBarrier

\section{Discussion and Limitations}
\label{sec:discussion_limitations}

GenPuzzle is intended to complement, rather than replace, existing image-generation and multimodal-reasoning benchmarks. Its focus is narrower but more executable: the model must preserve a concrete visual state, infer a rule-governed solution, and render that solution as an image. This design makes failures easy to diagnose. A low score can usually be traced to an invalid inference, an imprecise visual edit, corrupted input content, or a mismatch between the instruction and the generated output.

Our judge-selection experiment also clarifies the role of automated evaluation. Gemini 3.1 Pro agrees well with human evaluation at the aggregate level, reaching a Spearman correlation of 0.849 and a Pearson correlation of 0.912 over generation-model--track scores. This level of agreement is sufficient for scalable model ranking and error analysis, but it does not justify replacing human evaluation entirely. Ambiguous outputs, small symbols, thin paths, and partially correct visual constructions still require human review, and the benchmark explicitly routes such cases to manual inspection.

Thus, GenPuzzle uses MLLM evaluation as a high-throughput measurement layer rather than an oracle. The automatic judge provides consistent aggregate scores, enables full-benchmark comparison across generators, and reduces the amount of manual scoring needed for routine evaluation. Human evaluation remains the calibration and adjudication layer: it is used to choose the judge, validate the scoring direction, and resolve samples where the visual evidence is insufficient for a reliable automatic decision.

The benchmark also has limitations. Some tracks allow multiple valid solutions, so the reference image cannot be treated as the only target. We address this with task-specific rubrics and judge selection against human reference scores, but automatic judges may still misread small symbols, thin paths, or ambiguous geometric marks. GenPuzzle therefore reports aggregate trends rather than claiming fully reliable per-instance scoring. Future versions can expand the task families, add more human-labeled edge cases, and include interaction traces or edit histories to separate planning failures from rendering failures more precisely.

\section{Conclusion}
\label{sec:conclusion}

GenPuzzle evaluates visual reasoning through image generation: a model must solve a visual puzzle and render the solution while preserving the input state. The benchmark contains 2,005 problems across 12 balanced tracks and uses task-specific evaluation, combining grid transcription and programmatic checks with validated MLLM rubrics. Experiments show that frontier generators remain far from reliable visual problem solving: the best Macro Overall score is 40.57, with severe failures in matchsticks, mazes, geometry, and state preservation. GenPuzzle offers a testbed for executable, verifiable visual generation.

\bibliography{aaai2027}

\end{document}